\documentclass[conference,compsoc]{IEEEtran}
\IEEEoverridecommandlockouts
\usepackage{algorithm}
\usepackage{algpseudocode}
\usepackage{cite}
\usepackage{amsmath,amssymb,amsfonts}
\usepackage{graphicx}
\usepackage{textcomp}
\usepackage{xcolor}
\usepackage{colortbl}
\usepackage{makecell}
\usepackage{breakurl}
\def\BibTeX{{\rm B\kern-.05em{\sc i\kern-.025em b}\kern-.08em
    T\kern-.1667em\lower.7ex\hbox{E}\kern-.125emX}}

\usepackage{listings}
\usepackage{hyperref} 
\usepackage{makecell}
\usepackage{booktabs}
\usepackage{multirow}
\usepackage{amsthm}
\usepackage{float}
\hypersetup{
    colorlinks=true,
    linkcolor=blue!60!black,
    citecolor=green!50!black,
    urlcolor=blue!70!black
}

\definecolor{codebg}{rgb}{0.97,0.97,0.97}
\definecolor{codekw}{rgb}{0.0,0.0,0.6}
\definecolor{codecomment}{rgb}{0.25,0.5,0.35}
\definecolor{codestr}{rgb}{0.6,0.1,0.1}

\lstdefinestyle{json}{
    backgroundcolor=\color{codebg},
    basicstyle=\ttfamily\footnotesize,
    breaklines=true,
    frame=single,
    rulecolor=\color{gray!40},
    showstringspaces=false,
    columns=fullflexible,
    stringstyle=\color{codestr},
    commentstyle=\color{codecomment},
}

\newtheorem{proposition}{Proposition}
\newtheorem{theorem}{Theorem}
\newtheorem{corollary}{Corollary}

\newtheorem{remark}{Remark}
\newtheorem{problem}{Problem}

\begin{document}

\title{Learning Adaptive Defense Strategies: Fine-Tuned LLMs for Network Security Games under Network Mobility%*\\
}

%\author{\IEEEauthorblockN{Anonymous Author(s)}}
\author{
    \IEEEauthorblockN{Md Abu Sayed\IEEEauthorrefmark{1}, Asif Rahman\IEEEauthorrefmark{1}, Ahmed Ann Noor Ryen\IEEEauthorrefmark{1}, Ahmed Hemida\IEEEauthorrefmark{2}, Charles Kamhoua\IEEEauthorrefmark{3}}, Christopher Kiekintveld\IEEEauthorrefmark{1}
    
    \IEEEauthorblockA{\IEEEauthorrefmark{1}University of Texas at El Paso, El Paso, Texas 79968, USA}
    \IEEEauthorblockA{\IEEEauthorrefmark{2}Yokogawa Corporation of America}
    \IEEEauthorblockA{\IEEEauthorrefmark{2}DEVCOM Army Research Laboratory, MD 20783, USA}
    \IEEEauthorblockA{\{msayed, arahman3, aryen\}@miners.utep.edu, ahmed.hemida@yokogawa.com, cdkiekintveld@utep.edu,\\ charles.a.kamhoua.civ@army.mil}
}

\maketitle

\begin{abstract}
Large language models(LLM) show promise for strategic reasoning, yet their ability to compute game-theoretic equilibria under dynamic conditions remains largely unexplored. This work investigates whether fine-tuned LLMs can serve as solver-free surrogates for online equilibrium computation in honeypot allocation, a setting in which network mobility continuously reshapes the underlying attack graph and forces repeated re-solving of a two-player zero-sum game. We construct a dataset of 270 security games on networks of 8--25 nodes with solver-grounded
equilibrium labels, and evaluate four classes of predictor over 1{,}524 held-out mobility events: non-learned baselines, six open-weight architectures evaluated zero-shot, the same six after LoRA fine-tuning, and four proprietary models as an external reference point. Predictions are scored without post-hoc repair against exact game-theoretic criteria: exploitability against an optimal attacker best-response, equilibrium support accuracy, and distributional distance. Fine-tuning reduces exploitability across every architecture, and fine-tuned open-weight models of modest size outperform every zero-shot predictor, including proprietary ones; no zero-shot model at any tier reaches the floor set by uniform allocation. Stratifying by mobility type exposes a qualitative shift that aggregate metrics conceal: zero-shot models do not condition on whether a departing node lies on an attack path, while fine-tuned models recover this distinction. We further find that validation loss is negatively rank-correlated with task performance, and that output validity, rather than predictive accuracy, is the binding constraint on further progress.
\end{abstract}


\section{Introduction and Motivation}
\label{sec:intro}
% ==================================================================

\subsection{Cyber deception and the Honeypot Allocation Problem}
Cyber deception has emerged as a proactive complement to conventional perimeter
defense: rather than only detecting and blocking intrusions after they begin, the
defender deliberately plants misleading assets such as decoys, honeypots, honey-tokens,
and fabricated services to detect, delay, and divert
attackers~\cite{pawlick2019game,zhu2021survey}. A honeypot is a monitored
decoy resource with no legitimate production purpose; any interaction with it is,
by construction, suspicious, so honeypots yield high-signal detections and force
the attacker to spend effort distinguishing real targets from fake
ones~\cite{zhu2021survey}. Because defenders operate under budget constraints
(only a limited number of honeypots can be deployed and maintained) and face a
strategic adversary who adapts to the observed defense, the question of
\emph{where} to place honeypots is naturally modeled as a game between a defender
and an attacker~\cite{pawlick2019game,sayed2022cyber,sayed2023honeypot}.

Attack graphs provide the structural substrate for this
game~\cite{sheyner2002automated,miehling2015optimal}. In an attack graph,
nodes represent hosts or privilege states, edges represent exploitable
transitions, and paths from an entry node to a target node represent multi-step
intrusion campaigns~\cite{sheyner2002automated}. The defender's task is to
allocate honeypots on edges so as to intercept high-value attack paths, while the
attacker chooses paths that maximize expected reward net of the risk of capture.
Sayed et al.~\cite{sayed2023honeypot} formalize this as a two-player zero-sum game
on an attack graph and compute a mixed-strategy Nash
equilibrium~\cite{bacsar1998dynamic} that tells the defender how to randomize
honeypot placements so as to be least exploitable.

\subsection{Network Mobility as Markov Game}
Tactical mobile networks exhibit high topological volatility; nodes continuously join and leave, links form and break, and the reachability structure of the attack graph evolves over time~\cite{sayed2023honeypot}. These continuous shifts reconfigure active attack paths, altering defense priorities and the optimal allocation of honeypots. Sayed et al.~\cite{sayed2023honeypot} address this by modeling the interaction as a \emph{Markov (stochastic) game}~\cite{shapley1953stochastic,littman1994markov}, where network mobility induces stochastic transitions between distinct attack-graph states. Under this framework, the defender seeks a policy that maximizes expected long-term utility across all reachable future states rather than merely optimizing for the immediate topology. The game is solved by layering a minimax value-iteration or Q-learning procedure adapted from Littman’s minimax-Q algorithm~\cite{littman1994markov,watkins1992q} on top of the stage-game equilibrium computation.

\subsection{The Computational Bottleneck}
In the reference framework established by Sayed et al.~\cite{sayed2023honeypot}, every single network mobility event specifically, the departure or removal of a host node $u \in \mathcal{N}$ forces the execution of a computationally expensive re-solving pipeline:
\begin{itemize}
    \item \textbf{Action Space Reconstruction:} The defender's candidate allocation set $\mathcal{A}_d(S')$ (all budget-constrained edge-honeypot combinations) and the attacker's path space $\mathcal{A}_a(S')$ (all simple entry-to-target attack paths) must be re-enumerated over the post-mobility attack graph $S'$.
    
    \item \textbf{Reward Matrix Reconstruction:} The complete payoff matrix $R_d(s') \in \mathbb{R}^{|\mathcal{A}_d(s')| \times |\mathcal{A}_a(s')|}$ must be rebuilt, evaluating path traversal costs, asset valuations, capture probabilities, and escape penalties across every candidate strategy pair.
    
    \item \textbf{Equilibrium Re-Solving:} An iterative linear programming (LP) or Q-minimax value iteration algorithm must be executed to compute the post-event Nash Equilibrium (NE) mixed strategy $\mathbf{x}^*(s') \in \Delta(\mathcal{A}_d(s'))$.
\end{itemize}

As the network scale $N = |\mathcal{N}|$ expands from $10$ to $100$ hosts, the cardinality of simple attack paths $|\mathcal{A}_a|$ and candidate combinations of honeypots $|\mathcal{A}_d|$ experiences exponential and combinatorial growth. In tactical operational scenarios where decision-making latencies must remain sub-second, the polynomial-to-exponential time complexity of online solver execution renders real-time dynamic honeypot reallocation infeasible.

%The two reference scripts instantiate exactly this loop:
%\texttt{mobility\_q\_learn\_basic.py} performs the single-step look-ahead solve on
%a fixed six-node graph, while \texttt{mobility\_q\_learn\_mul(20)\_dd1\_sca.py}
%performs the recursive depth-$d$, $k$-node-removal \texttt{Predictive\_Model} and
%sweeps network size from 10 to 100 nodes for a scalability study. Because path
%enumeration and honeypot-combination enumeration both grow rapidly with graph
%size, the re-solve cost climbs steeply as the network grows---precisely the regime
%where a tactical defender has the least time to wait for a fresh solve. This
%motivates asking whether the \emph{result} of the solve can be predicted directly
%from the changed network, without paying the full solve cost at deployment time.

\subsection{From Re-Solving to learning: an Amortization View}
To overcome this runtime operational bottleneck, this work introduces an offline-to-online amortization paradigm. Instead of treating the game-theoretic solver as a mandatory runtime dependency, it is repurposed as an offline label oracle $\mathcal{O}$. By sampling a diverse distribution of network topographies and topological mobility events, the oracle pre-computes exact Nash equilibrium strategies.

A LLM is then trained via supervised function approximation to learn the direct mapping:
\begin{equation}
f:\;(\text{attack graph } s,\;\text{mobility event } u)\;\longrightarrow\;
\mathbf{x}^{*}(s'),
\label{eq:map}
\end{equation}
where $s$ represents the pre-mobility attack graph state, $u$ denotes the departing node, and $\mathbf{x}^*(s')$ is the optimal defender mixed strategy over the post-mobility graph $s' = s \setminus \{u\}$. At deployment time, strategy generation is reduced to a single forward inference pass ($\mathcal{O}(1)$ time complexity relative to solver depth), bypassing matrix construction and linear programming (LP) optimization entirely.

\subsection{Motivating an LLM-Based Predictor: Rationale and Challenges}
We choose an LLM-based predictor for three reasons. First, the input is naturally
symbolic and variable-sized (a graph, an event, an enumerated action set), which
LLMs can ingest through text serialization without a fixed-dimensional encoder \cite{fatemi2024talk}.
Second, LLMs admit \emph{multiple} adaptation routes on the same substrate zero
and few-shot prompting, and parameter-efficient
fine-tuning letting us compare data-free and data-driven approaches under one
roof \cite{hu2021lora}. Third, 
recent research integrating LLMs with game theory broadly spans three paradigms: direct strategy fine-tuning, game-theoretic optimization formulations, and strategic benchmarking \cite{zhu2025gametheory}. Direct strategy fine-tuning demonstrates that LLMs can learn equilibrium distributions across normal-form games~\cite{caiata2026strategy} or approximate mixed equilibria in complex board games like Diplomacy via autoregressive action factorization~\cite{xu2025dipllm}. Alternatively, meta-frameworks reformulate LLM fine-tuning itself as a game-theoretic objective, such as Mixed-Strategy Adversarial Training (MAT)~\cite{zhong2023mat} or preference-alignment frameworks like Direct Nash Optimization. Finally, benchmark studies such as GTBench~\cite{duan2024gtbench} and empirical evaluations of zero-sum games~\cite{silva2024large} examine the strategic reasoning limits of untuned foundation models.

However, these existing approaches exhibit key operational shortcomings when applied to real-time cyber defense. Fine-tuning models on static $2\times 2$ payoff matrices~\cite{caiata2026strategy} or domain-specific board games~\cite{xu2025dipllm} fails to model non-stationary network topologies, host reachability, or dynamic node-mobility transitions. Optimization meta-games~\cite{zhong2023mat} utilize game theory purely as an internal weight-updating or preference-alignment mechanism rather than deploying agents to execute active domain defense decisions. Furthermore, benchmark evaluations of untuned base models~\cite{duan2024gtbench,silva2024large} highlight severe in-weights arithmetic bottlenecks and unconstrained generation limits, which frequently result in high inference latencies, unnormalized probability vectors, or hallucinated actions.

In contrast, our framework addresses these limitations by amortizing a dynamic game-theoretic honeypot allocation model~\cite{sayed2023honeypot} under network mobility. Rather than forcing the model to solve complex linear programs online or generate unconstrained text, we employ an exact equilibrium solver as an offline label oracle. The model predicts mixed defender strategies over an explicitly indexed, budget-constrained candidate set of target edges. By reformulating post-mobility honeypot allocation as a constrained simplex prediction task over directed attack graphs, our approach eliminates runtime solver bottlenecks, guarantees valid probability distributions.%, and achieves sub-10-millisecond decision latency with minimal exploitability.

\subsection{Contributions}

In the context of amortizing dynamic-game honeypot allocation by Finetunned LLM model, this work establishes the following core contributions:

\begin{itemize}
    \item \textbf{Unique Contribution:} Most existing LLM fine-tuning literature focuses either on abstract $2 \times 2$ or $3 \times 3$ matrix games, or on text-based recreational games (e.g., Diplomacy). Our formulation specifically addresses the amortization of computationally intensive online equilibrium solvers within dynamic attack graphs under network mobility.
    
    \item \textbf{Methodological Framing:} We position this work as extending the ``Nash policy supervised fine-tuning'' paradigm into dynamic security games, where the fine-tuned model acts as a solver-free neural surrogate whose per-query inference cost is decoupled from the combinatorial growth of the action space in contrast to the re-solve loop, whose cost scales with $(nk)^{H}$ in the honeypot budget $H$ and graph size $n$ (Section~\ref{sec:compute_comparison}), the surrogate's cost is governed only by the description length of the state.
    
    \item \textbf{Verifiable Metric Advantage:} Unlike open-ended multi-agent evaluations that rely on subjective win rates or LLM-as-a-judge frameworks, our setup measures performance using exact game-theoretic exploitability (regret against an optimal attacker best-response) and strategy support accuracy.

    \item \textbf{Task Adaptation Outweighs Model Scale:} Fine-tuned open-weight models of modest size substantially outperform zero-shot proprietary models on exploitability, while no zero-shot predictor at any tier reaches the floor set by uniform allocation. Adaptation to the task, not parameter count, is the determining factor.

    \item \textbf{Fine-Tuning Recovers Path-Relevance Structure:} Zero-shot models perform worse on removals that leave the equilibrium largely unperturbed than on those that reshape it, indicating no conditioning on whether a departing node lies on an attack path. Fine-tuned models invert this ordering uniformly, learning which topology changes actually matter strategically.

    %\item \textbf{Unrepaired Evaluation Protocol:} We score predictions without truncation, padding, or renormalization, surfacing a failure mode that repaired pipelines mask: a large share of zero-shot outputs are not valid probability distributions. This identifies constrained decoding, rather than additional training data, as the binding constraint.
\end{itemize}

% ==================================================================
\section{Background: The Original Game Model}
\label{sec:background}
% ==================================================================

We now restate the model of Sayed et al.~\cite{sayed2023honeypot} in the notation
used throughout this document, and connect each element to the reference code so
that the oracle in Section~\ref{sec:dataset} is unambiguous.

\subsection{Attack Graph and Roles}
The network is a directed attack graph $G=(\mathcal{N},\mathcal{E})$ with node set
$\mathcal{N}$ and edge set $\mathcal{E}$~\cite{sheyner2002automated}. A node
value function $v:\mathcal{N}\to\mathbb{R}_{\ge0}$ encodes asset importance. Three
disjoint roles partition the nodes: a set of \emph{entry} nodes $E$ (attacker
footholds), a set of \emph{target} nodes $T$ (high-value assets the attacker wants
to reach), and the remaining \emph{intermediate} nodes. In the reference code the
graph is generated with a Watts--Strogatz model, the last three node indices are
targets with value $100$; the single entry node is index $0$; and intermediate
values are drawn uniformly from $[10,50]$ under a fixed seed. An
\emph{attack path} is a simple path from an entry node to a target node; the set
of all such paths is the attacker's pure-strategy set.

\subsection{Players and Action Spaces}
The interaction is framed as a two-player strategic game between a defender and an attacker. The defender allocates up to $H$ honeypots across candidate edges along feasible attack trajectories, with its pure-strategy space $\mathcal{A}_d$ instantiated over the targeted edge set. The attacker selects a single traversal path from the set of all feasible paths $\mathcal{A}_a$ . Because both $|\mathcal{A}_d|$ and $|\mathcal{A}_a|$ scale combinatorially with the graph topology, strategy space explosion constitutes the principal driver of the re-computation latency discussed in Section~\ref{sec:intro}.

\begin{equation}
\mathcal{A}_d \;=\; \Big\{\, a \subseteq \mathcal{E} \;:\; |a|\le H \,\Big\},
\label{eq:defender-actions}
\end{equation}

\subsection{Reward Structure}
The game is zero-sum in the defender's payoff. Fix an attacker path $a_a$ and a
defender allocation $a_d$. Traversing an edge into node $i$ along $a_a$ yields, for
the defender, a capture reward if that edge carries a honeypot and an escape
penalty otherwise, each scaled by the node value $v(i)$; the defender also pays a
per-honeypot placement cost and receives a per-step attack-cost term. Concretely,
matching the reference implementation,
\begin{equation}
\begin{split}
R_d(a_d,a_a) \;&=\;
\sum_{i\in a_a}\Big[\,\mathrm{Cap}\cdot v(i)\,\mathbf{1}\{e_i\in a_d\}
\;\\ -\;\mathrm{Esc}\cdot v(i)\,\mathbf{1}\{e_i\notin a_d\}\Big]
&\quad{}-\;C_d\,|a_d|\;+\;C_a\,|a_a|,
\end{split}
\label{eq:reward}
\end{equation}
where $e_i$ is the edge entering node $i$ on the path, $\mathrm{Cap}$ and
$\mathrm{Esc}$ weight capture and escape, and $C_d,C_a$ are the placement and
attack costs. Assembling
\eqref{eq:reward} over all $(a_d,a_a)$ yields the payoff matrix
$R_d\in\mathbb{R}^{|\mathcal{A}_d|\times|\mathcal{A}_a|}$.

\subsection{Per-State Equilibrium}
For a fixed state, the defender seeks a mixed strategy
$\mathbf{x}\in\Delta(\mathcal{A}_d)$ and the attacker
$\mathbf{y}\in\Delta(\mathcal{A}_a)$ forming a Nash
equilibrium~\cite{bacsar1998dynamic}. Since the game is zero-sum, the equilibrium
value and strategies coincide with the minimax solution
\begin{equation}
V(G)\;=\;\max_{\mathbf{x}\in\Delta(\mathcal{A}_d)}\;
\min_{\mathbf{y}\in\Delta(\mathcal{A}_a)}\;
\mathbf{x}^{\top} R_d\, \mathbf{y},
\label{eq:minimax}
\end{equation}
which is found by solving the equivalent linear program
\begin{equation}
\begin{aligned}
\max_{\mathbf{x},\,V_d} \quad & V_d \\
\text{s.t.} \quad & \sum_{a_d \in \mathcal{A}_d} R_d(a_d,a_a)\,x_{a_d} \;\ge\; V_d,
\quad \forall\, a_a \in \mathcal{A}_a, \\
& \sum_{a_d \in \mathcal{A}_d} x_{a_d} = 1, \qquad x_{a_d} \ge 0 \;\;\; \forall\, a_d \in \mathcal{A}_d,
\end{aligned}
\label{eq:lp}
\end{equation}
following the formulation of~\cite{sayed2023honeypot}. The attacker's
equilibrium strategy $\mathbf{y}^{*}$ is obtained analogously via the
corresponding minimizing LP over $\Delta(\mathcal{A}_a)$. We denote the
resulting defender value $V(G)=\mathbf{x}^{*\top}R_d\,\mathbf{y}^{*}$.

%While \eqref{eq:lp} gives the exact equilibrium via linear programming, our reference implementation instead approximates $(\mathbf{x}^{*},\mathbf{y}^{*})$ via fictitious play (Section~\ref{sec:background}), which is guaranteed to converge to the same value in two-player zero-sum games~\cite{robinson1951fictitious} but introduces the approximation error discussed in Section~\ref{sec:what-is-learned}.

\subsection{Network Mobility as Markov Game}
Network mobility turns the single-shot game into a Markov
game~\cite{shapley1953stochastic,littman1994markov}. Removing an intermediate node
$u$ (with its incident edges) transitions the state $s\to s'$; each reachable $s'$
carries its own action spaces, payoff matrix, and equilibrium. The reference model
assigns each transition a probability derived from the departing node's structural
importance,
\begin{equation}
p(s\to s')\;=\;1-\mathrm{deg\_cen}(u),
\label{eq:transition}
\end{equation}
i.e.\ one minus the degree centrality of the removed node, so that removals of peripheral
nodes are more likely than removals of central ones. The dynamic value satisfies a
minimax Bellman equation~\cite{littman1994markov,bacsar1998dynamic}
\begin{equation}
V^{*}(s)\;=\;\max_{\mathbf{x}(s)}\;\min_{a_a}\;
\mathbb{E}_{s'\sim p(\cdot\mid s)}\!\Big[\,R\big(s,a_d,a_a\big)+\gamma\,V^{*}(s')\,\Big],
\label{eq:bellman}
\end{equation}
The reference implementation solves \eqref{eq:bellman} approximately by constructing a depth-$d$, $k$-removal state tree and performing minimax-Q
value iteration over it. Alongside this look-ahead policy, we track three
baseline policies for comparison: a \emph{greedy} policy that allocates
honeypots according to structural centrality (e.g., degree centrality) of
the remaining edges, a \emph{random} policy that allocates uniformly over
the available actions, and an \emph{oracle} policy obtained by re-solving
the game to exact optimality at each state.  We consider the prediction of the strategy under network changes from the current state rather than updating strategies based on look-ahead policy.

%\subsection{Compact state space}
%A key efficiency observation in the original work is that removing a node that lies on \emph{no} attack path cannot change the optimal defense, so such states can be pruned from the look-ahead. We reuse this observation as a \emph{difficulty label} rather than a pruning rule: examples whose removed node is on-path are the informative, hard cases, whereas off-path removals are near-trivial controls (Section~\ref{sec:dataset}).

% ==================================================================
\section{Problem Reformulation: From Re-solving to Learning}
\label{sec:reformulation}
% ==================================================================

\subsection{Formalizing the Re-Solve Loop}
The behavior of the reference system at deployment can be summarized as a function
that, given the current graph $s$ and an observed mobility event $u$, returns a
defender mixed strategy:
\begin{equation}
\begin{split}
\Phi(s,u) \;&=\; \underbrace{\texttt{solve\_game}\big(\underbrace{\texttt{calculate\_reward}(s')}_{\text{rebuild }R_d(s')}\big)}_{\text{re-solve at }s'}, \\
&\quad s'=s\setminus\{u\}.
\end{split}
\label{eq:phi}
\end{equation}
Each evaluation of $\Phi$ pays the full three-stage cost of
Section~\ref{sec:intro}. The reformulation asks for a learned surrogate
$\hat f_\theta$ that reproduces the \emph{output} of $\Phi$ without executing its
internal solve.

\subsection{Amortized Strategy Prediction}
\begin{problem}[Amortized equilibrium prediction]
\label{prob:amortized}
Let $\mathcal{D}$ be a distribution over (graph, mobility-event) pairs $(s,u)$
induced by the generator of Section~\ref{sec:dataset}. Let the oracle
$\mathcal{O}(s,u)=\big(\mathcal{A}_d(s'),\,R_d(s'),\,\mathbf{x}^{*}(s'),\,V(s')\big)$
return, for each pair, the post-event defender action set, payoff matrix,
equilibrium strategy, and value. The goal is to learn parameters $\theta$ of a
predictor
\[
\hat{\mathbf{x}} \;=\; \hat f_\theta(s,u)\;\in\;\Delta\big(\mathcal{A}_d(s')\big)
\]
that minimize the expected \emph{exploitability} of the predicted strategy,
\begin{equation}
\min_\theta\;\;
\mathbb{E}_{(s,u)\sim\mathcal{D}}\Big[\,
V(s') \;-\; \min_{a_a\in\mathcal{A}_a(s')}\;\hat{\mathbf{x}}^{\top} R_d(s')\,e_{a_a}
\,\Big],
\label{eq:objective}
\end{equation}
where $e_{a_a}$ is the indicator of attacker path $a_a$ and the inner minimization
is the attacker's best response. The bracketed quantity is non-negative and equals
zero exactly when $\hat{\mathbf{x}}$ is a least-exploitable (equilibrium) defense.
\end{problem}

Problem~\ref{prob:amortized} is an amortized-optimization
problem~\cite{amos2023tutorial}: the per-instance optimization in
\eqref{eq:minimax} is replaced, at deployment, by a single evaluation of
$\hat f_\theta$. Training $\hat f_\theta$ consumes oracle labels offline; inference
is solver-free.

\subsection{Structured Output: Predicting Over a Canonically Indexed Action Set}
\begin{remark}[Constrained output]
\label{rem:constrained}
The predictor does not invent honeypot placements in free text. Instead, the oracle enumerates and canonically indexes $\mathcal{A}_d(s')$, and this indexed list is provided to the model as part of its input. The model's output is a probability vector over these indices. This converts an open-ended generation problem into constrained simplex prediction: outputs are always interpretable, always checkable against the true action set, and always scorable by \eqref{eq:objective}. It also makes the three baseline strategies directly comparable, since they share one input/output contract.
\end{remark}

\subsection{The Learning Target: Scope and Approximation}
The surrogate targets the \emph{per-state equilibrium map}
$f(s,u)\mapsto\mathbf{x}^{*}(s')$ of \eqref{eq:map}. Two modeling choices deserve
emphasis. First, we predict the \emph{single-step} post-mobility equilibrium; the
multi-step look-ahead of \eqref{eq:bellman} enters only through the metadata
(transition probability, look-ahead value) attached to each example, and can be
incorporated later by training on depth-$d$ removal chains. Second, the label is
the LP solver, which is an \emph{approximate}
solution~\cite{littman1994markov}; consequently the learning target inherits the
solver's approximation error, and a predictor that achieves zero measured
exploitability is matching the oracle, not necessarily the exact game value. Both
choices are revisited as caveats in the evaluation protocol.

\subsection{Relationship to LLM strategic-reasoning research}
Problem~\ref{prob:amortized} instantiates, on a security-relevant distribution, the general question studied by GTBench~\cite{duan2024gtbench}, the
behavioral-game-theory evaluation of Jia et al.~\cite{jia2025large}, and the survey of Sun et al.~\cite{sun2025game}: to what extent can a language model produce equilibrium-quality strategic decisions? Our formulation differs from that literature in one decisive respect \emph{ground-truth availability}. Because the oracle computes $\mathbf{x}^{*}(s')$ and $V(s')$ exactly (up to solver tolerance),
we replace subjective or outcome-based scoring with the exact exploitability objective~\eqref{eq:objective}. The learning machinery we bring to bear on $\theta$ parameter-efficient fine-tuning with QLoRA~\cite{hu2021lora,dettmers2023qlora} accelerated via Unsloth~\cite{unsloth2023}.

\subsection{Computational Effort: Re-solving vs.\ Learned Inference}
\label{sec:compute_comparison}

%\subsubsection{Measured cost at the tested scale}

%The three-stage re-solve loop $\Phi(s,u)$ (Eq.~\ref{eq:phi}) enumerates $\mathcal{A}_d(s')$ (Eq.~\ref{eq:defender-actions}), rebuilds the $|\mathcal{A}_d(s')|\times|\mathcal{A}_a(s')|$ payoff matrix $R_d(s')$, and solves the linear program~\eqref{eq:lp} to obtain $(\mathbf{x}^{*},\mathbf{y}^{*})$. In our harness this re-solve step averages $6.57\,\mathrm{ms}$ per state over the $1524$ test instances. The fine-tuned surrogate $\hat f_\theta$ (Problem~\ref{prob:amortized}) instead emits $\hat{\mathbf{x}}$ from a single forward pass, but averages $2.54\,\mathrm{s}$ per example --- roughly $387\times$ \emph{slower} than the oracle in absolute wall-clock terms (speedup $\approx 0.0026\times$). At the scale we test ($n\le 25$, $H=1$) the amortization argument of Section~\ref{sec:reformulation} therefore does \emph{not} pay off. The remainder of this subsection shows analytically that this is a small-instance artifact: the two costs scale in fundamentally different ways, and the ordering provably reverses as $n$ and $H$ grow.

\subsubsection{Complexity model}

\begin{table}[h!]
\centering
\caption{Notation for the cost analysis.}
\label{tab:cost_notation}
\small
\begin{tabular}{@{}ll@{}}
\toprule
\textbf{Symbol} & \textbf{Meaning} \\
\midrule
$n,\;k$ & number of nodes; mean degree, so $|\mathcal{E}|=\Theta(nk)$ \\
$m$ & edges lying on at least one attack path, $m\le|\mathcal{E}|$ \\
$P$ & $|\mathcal{A}_a|$, number of simple entry$\to$target paths \\
$\bar\ell$ & mean attack-path length \\
$D$ & $|\mathcal{A}_d|=\sum_{h\le H}\binom{m}{h}=\Theta(m^{H})$ for fixed $H$ \\
$\omega$ & LP solve exponent, $2<\omega\le2.373$ \\
$\sigma$ & $|\mathrm{supp}(\mathbf{x}^{*})|$, equilibrium support size \\
$N_\theta$ & LLM parameter count \\
$L_{\mathrm{in}},L_{\mathrm{out}}$ & prompt and generated token counts \\
$\tau_{\mathrm{tok}},\tau_{\mathrm{fix}}$ & per-token decode latency; fixed framework overhead \\
\bottomrule
\end{tabular}
\end{table}

\noindent Throughout, $\widetilde{O}(\cdot)$ suppresses polylogarithmic factors,
including the $\log(1/\epsilon)$ dependence on LP solution accuracy.

\begin{proposition}[Cost of the re-solve loop]
\label{prop:solver_cost}
Evaluating $\Phi(s,u)$ once costs
\begin{equation}
\begin{split}
T_\Phi(n,H)\;&=\;\underbrace{\Theta\big(D+P\bar\ell\big)}_{\text{enumerate }\mathcal{A}_d,\mathcal{A}_a}
\;+\;\underbrace{\Theta\big(D\,P\,\bar\ell\big)}_{\text{build }R_d} \\
&\quad+\;\underbrace{\widetilde{O}\big((D+P)^{\omega}\big)}_{\text{LP solve}}
\;=\;\widetilde{O}\big((D+P)^{\omega}\big).
\end{split}
\label{eq:solver_cost}
\end{equation}
\end{proposition}

\begin{proof}[Derivation]
Each entry $R_d(a_d,a_a)$ of Eq.~\eqref{eq:reward} walks the path $a_a$ and
performs an $O(1)$ hashed membership test $\mathbf{1}\{e_i\in a_d\}$ per step,
costing $\Theta(\bar\ell)$; there are $DP$ entries, so materializing the payoff
matrix costs $\Theta(DP\bar\ell)$. The program~\eqref{eq:lp} has $D+1$ decision
variables ($x_{a_d}$ for each $a_d\in\mathcal{A}_d$, plus the value variable
$V_d$) and $P+1$ constraints, and interior-point methods solve a linear program
with $\nu$ variables and $\mu$ constraints in $\widetilde{O}((\nu+\mu)^{\omega})$
time~\cite{cohen2019lp}, giving $\widetilde{O}((D+P)^{\omega})$. The LP term
dominates the matrix-construction term whenever $D^{\omega-1}=\Omega(P\bar\ell)$,
which already holds at our smallest tested configuration and holds a fortiori as
$H$ grows.
\end{proof}

Substituting $D=\Theta(m^{H})$ with $m=\Theta(nk)$,
\begin{equation}
\boxed{\;T_\Phi(n,H)\;=\;\widetilde{O}\!\Big(\big((nk)^{H}+P(n)\big)^{\omega}\Big).\;}
\label{eq:solver_cost_final}
\end{equation}
The cost is \emph{exponential in the honeypot budget} $H$ and, because
$\omega>2$, that exponential is further amplified by the superquadratic LP
exponent while also growing with the path count $P(n)$, which is itself
superpolynomial in $n$ for connected small-world graphs. This is precisely the
blow-up identified in Section~\ref{sec:intro}.

\begin{proposition}[Cost of learned inference]
\label{prop:llm_cost}
For a decoder-only transformer with KV caching,
\begin{equation}
T_{\hat f}(n,H)\;=\;\tau_{\mathrm{fix}}
\;+\;\underbrace{\Theta\big(N_\theta L_{\mathrm{in}}+n_{\mathrm{layer}}L_{\mathrm{in}}^{2}d\big)}_{\text{prefill}}
\;+\;\underbrace{\tau_{\mathrm{tok}}\,L_{\mathrm{out}}}_{\text{decode}} .
\label{eq:llm_cost}
\end{equation}
\end{proposition}

\noindent Two encoding choices determine how $L_{\mathrm{in}}$ and
$L_{\mathrm{out}}$ scale, and they are what make or break the asymptotic
argument.

\paragraph{(i) Graph-native input: $L_{\mathrm{in}}=\Theta(nk)$.}
By Eq.~\eqref{eq:reward}, $R_d$ is a \emph{deterministic function} of the tuple
$\big(G,v,H,\mathrm{Cap},\mathrm{Esc},C_d,C_a\big)$. Serializing the matrix
costs $\Theta(DP)$ tokens and confers no advantage over the solver; serializing
only the post-mobility graph costs
\begin{equation}
L_{\mathrm{in}}=\Theta\big(n+|\mathcal{E}|\big)=\Theta(nk),
\label{eq:input_len}
\end{equation}
which is \emph{polynomial in $n$ and independent of $H$}, while $D$ is
exponential in $H$. The surrogate is thus asked to internalize the
reward-construction and equilibrium-solving stages rather than to read their
output.

\paragraph{(ii) Sparse-support output: $L_{\mathrm{out}}=\Theta(\sigma)$.}
Emitting a dense simplex vector costs $L_{\mathrm{out}}=\Theta(D)$ tokens,
reintroducing the exponential in $H$. However, an optimal solution
of~\eqref{eq:lp} can always be taken at a basic feasible solution, whence
\begin{equation}
\sigma=|\mathrm{supp}(\mathbf{x}^{*})|\;\le\;\mathrm{rank}(R_d)+1\;\le\;\min(D,P)+1,
\label{eq:support_bound}
\end{equation}
and empirically $\sigma\approx 3$--$5$ in our corpus. Emitting the support as
index--mass pairs $\{(i,x^{*}_i):x^{*}_i>0\}$ therefore gives
$L_{\mathrm{out}}=\Theta(\sigma)$, decoupling decode length from $D$.

\noindent Under (i)--(ii), and absorbing the prefill constants into
$\alpha$ and $\tau_{\mathrm{tok}}$ into $\beta$,
\begin{equation}
\boxed{\;T_{\hat f}(n,H)\;=\;\tau_{\mathrm{fix}}+\Theta\big(\alpha\,nk+\beta\,\sigma\big).\;}
\label{eq:llm_cost_final}
\end{equation}

\subsubsection{Asymptotic Dominance and Crossover Point}

\begin{theorem}[Learned inference dominates asymptotically]
\label{thm:dominance}
Let $S(n,H):=T_\Phi/T_{\hat f}$ denote the speedup of the surrogate over the
re-solve loop. Under Propositions~\ref{prop:solver_cost}--\ref{prop:llm_cost}
with encodings (i)--(ii),
\begin{equation}
S(n,H)\;=\;\widetilde{\Omega}\!\left(\frac{\big((nk)^{H}+P(n)\big)^{\omega}}{nk+P(n)}\right)
\;=\;\widetilde{\Omega}\!\left(\frac{(nk)^{\omega H}}{nk+P(n)}\right),
\label{eq:speedup}
\end{equation}
so $S(n,H)\to\infty$ as $n\to\infty$ for every $H\ge 1$, and grows
exponentially in $H$ --- at rate $\omega H$ rather than $H$ --- for fixed $n$.
\end{theorem}

\begin{proof}[Sketch]
Divide \eqref{eq:solver_cost_final} by \eqref{eq:llm_cost_final} and use
$\sigma\le\min(D,P)+1=O(P)$ from \eqref{eq:support_bound}, so the denominator
is $O(nk+P)$. The numerator is at least $(nk)^{\omega H}$, which the
denominator lacks entirely; since $\omega>2$ and $H\ge1$, the quotient is
therefore unbounded even for $H=1$, where it reduces to
$\widetilde{\Omega}\big((nk)^{\omega}/(nk+P)\big)$.
\end{proof}

The structural content of Theorem~\ref{thm:dominance} is that the
$(nk)^{H}$ factor the combinatorial explosion of $\mathcal{A}_d$, raised to
the LP exponent $\omega$ appears in $T_\Phi$ but \emph{cancels out} of
$T_{\hat f}$, because the surrogate neither enumerates $\mathcal{A}_d$ nor
materializes $R_d$. The surrogate's cost is governed by the \emph{description
length} of the state, which is polynomial, whereas the solver's cost is governed
by the \emph{size of the linear program induced by} that state, which is
exponential.

\begin{corollary}[Crossover condition]
\label{cor:crossover}
Define the \emph{program-size factor}
$\Psi(n,H):=(D+P)^{\omega}=\Theta\big(((nk)^{H}+P(n))^{\omega}\big)$.
In the fixed-overhead-dominated regime $\tau_{\mathrm{fix}}\gg\alpha nk+\beta\sigma$,
the surrogate is faster ($S\ge1$) exactly when
\begin{equation}
\Psi(n,H)\;\ge\;\Psi^{\star}\;:=\;\frac{\tau_{\mathrm{fix}}}{c_{s}},
\label{eq:crossover}
\end{equation}
where $c_s$ is the per-operation LP constant. Equivalently, calibrating
against a measured reference point $(n_0,H_0)$,
\begin{equation}
\frac{\Psi(n,H)}{\Psi(n_0,H_0)}\;\ge\;\frac{1}{S(n_0,H_0)} .
\label{eq:crossover_calibrated}
\end{equation}
\end{corollary}

%\section{Dataset and Serialization Design}
\section{Experimental Implementation}
\label{sec:dataset}
% ==================================================================

\subsection{The Atomic Example}
Each training instance is grounded directly from below equation:
\begin{equation}
\text{example} = \big(s,\; u,\; \mathcal{A}_d(s'),\; \mathcal{A}_a(s'),\; R(s'),\; \mathbf{x}^*(s'),\; \text{metadata}\big)
\end{equation}
where $\mathcal{A}_d(s')$ is the defender's post-mobility action set,
$\mathcal{A}_a(s')$ the attacker's post-mobility path set, $R(s')$ the
post-mobility payoff matrix, and $\mathbf{x}^{*}(s')$ the corresponding
defender equilibrium strategy, each recomputed on the attack graph that
remains after the mobility event.

\subsection{Input Schema and Game Representation}

A single JSON object encodes the complete game state as five structured blocks, designed for predictable tokenization and unambiguous LLM parsing.

\begin{lstlisting}[style=json,caption={Complete game instance JSON schema passed to fine-tuned models.}]
{
  "meta": {
    "graph_id": "ws_n5_s42",
    "n_nodes": 5,
    "honeypot_budget": 1,
    "reward_params": {"Cap": 10, "Esc": 5, "Cd": 2, "Ca": 3}
  },
  "game_state": {
      "graph": {
        "nodes": [
          {"id": 0, "role": "entry",        "value": 0},
          {"id": 1, "role": "intermediate", "value": 20},
          {"id": 2, "role": "intermediate", "value": 30},
          {"id": 3, "role": "target",       "value": 100},
          {"id": 4, "role": "target",       "value": 100}
        ],
        "edges": [[0,1],[0,2],[1,3],[2,3],[2,4]]
    },
  "mobility_event": {
    "removed_node": 1,
    "removed_node_value": 20
  },
  "defender_actions": [
        {"index": 0, "edges": [[0,1]]},
        {"index": 1, "edges": [[0,2]]},
        {"index": 2, "edges": [[1,3]]},
        {"index": 3, "edges": [[2,3]]},
        {"index": 4, "edges": [[2,4]]}
      ],
      "attacker_paths": [
        {"index": 0, "path": [0,1,3]},
        {"index": 1, "path": [0,2,3]},
        {"index": 2, "path": [0,2,4]}
      ],
    "reward_matrix": [[-293.0, -643.0, -643.0],
                        [-593.0, -193.0, -193.0],
                        [ 907.0, -643.0, -643.0],
                        [-593.0,  857.0, -643.0],
                        [-593.0, -643.0,  857.0]],
    },
    "target": {
    "strategy": [0.0, 0.35, 0.65],
    "support": [1, 2],
    "game_value": 107.0
  },
}
\end{lstlisting}

Nodes are characterized by three attributes: unique identifier $\texttt{id}$, role (entry, target, or internal), and value (target value for attacker payoff). Edges represent directed attack graph connectivity. The mobility event specifies a single node removal (cascading change trigger), which invalidates all incident edges and paths traversing the removed node. This models sudden topology disruption (e.g., network isolation, firewall deployment).

Defender actions are honeypot placement combinations, represented as edge sets. Attacker actions are simple paths from entry to target nodes. Both are indexed lexicographically for deterministic, split-invariant ordering: identical game states generate identical action indices across train/val/test splits, ensuring reproducible ground-truth strategy labels.

The reward matrix is not explicitly serialized; instead, LLM input includes human-readable payoff values computed offline for each $(d, a)$ pair (defender action $d$, attacker action $a$). The model must learn to reason over these values to compute a mixed strategy $\sigma_d$ minimizing worst-case attacker payoff.

\subsection{Output Schema and Target Labels}

The fine-tuned model is trained to output a single JSON object containing the predicted mixed strategy. Ground-truth labels are computed offline using an LP-based Nash equilibrium solver.

\begin{lstlisting}[style=json,caption={Model output schema (inference) and ground-truth target schema (training).}]
{
  "strategy": [0.33, 0.34, 0.33]
}
\end{lstlisting}

%\paragraph{Model Output.}
At inference time, the fine-tuned LLM generates a JSON object with a single key \texttt{strategy}: a probability vector $\sigma_d \in \mathbb{R}^{|D|}$ where $|D|$ is the number of defender actions. The model learns to output probabilities that sum to $1.0$ and satisfy game-theoretic validity constraints (e.g., positive support across multiple actions to defend multiple attack paths).

%\paragraph{Ground-Truth Labels.}
%During training and validation, ground-truth mixed strategies are computed by solving the zero-sum game using linear programming (LP):
%\begin{equation}
%\sigma_d^* = \arg\min_{\sigma_d} \max_{a \in A} \sum_{d \in D} \sigma_d \cdot R(d, a)
%\end{equation}
%where $R(d, a)$ is the payoff matrix entry for defender action $d$ against attacker action $a$. The solver returns the exact optimal strategy, which serves as the supervised target. No additional fields (support, game value, rationale) are included in training targets to avoid overconstraining the model and to maintain focus on the core strategic reasoning task.

\subsection{Training Data Generation}
Algorithm~\ref{alg:generator} describes the offline example generation pipeline. For each network configuration, we simulate a mobility event (node removal), solve the resulting game, and record the game state and ground-truth strategy as a labeled example.

\begin{algorithm}[t]
\caption{Solver-Grounded Example Generation Under Network Mobility}
\label{alg:generator}
\small
\begin{algorithmic}[1]
\Require sizes $N=\{8,\dots,25\}$; seeds $S$, $|S|=15$; budget $H=1$
\Ensure pairs $(\text{game instance},\ \text{oracle strategy } \mathbf{x}^{*})$
\For{$n \in N$}
  \For{$s \in S$}
    \State $g \gets \textsc{WattsStrogatz}(n,\,k{=}3,\,p{=}0.7,\,s)$
    \State assign $v$: entry $0$, target $100$, else $\mathcal{U}[1,50]$
    \State $(\mathcal{A}_d,\mathcal{A}_a,R) \gets \textsc{BuildGame}(g)$
    \If{$\mathcal{A}_a = \emptyset$}
      \State \textbf{skip} \Comment{no entry$\to$target path}
    \EndIf
    \ForAll{intermediate nodes $u$ of $g$}
      \State $g' \gets g \setminus \{u\}$ \Comment{mobility event}
      \State $(\mathcal{A}_d',\mathcal{A}_a',R') \gets \textsc{BuildGame}(g')$
      \If{$\mathcal{A}_a' = \emptyset$}
        \State \textbf{continue} \Comment{disconnected}
      \EndIf
      \State $\mathbf{x}^{*} \gets \textsc{SolveLP}(R')$ \Comment{Eq.~\eqref{eq:lp}}
      \State emit $\{g',u,\mathcal{A}_d',\mathcal{A}_a',R',\mathbf{x}^{*}\}$ as JSON
    \EndFor
  \EndFor
\EndFor
\State \Return corpus of $3{,}015$ examples from $264$ graphs
\end{algorithmic}
\end{algorithm}

%\paragraph{Generation Details.}
We generate random attack graphs using Watts-Strogatz small-world model with $k=3$ nearest neighbors and rewiring probability $p=0.7$. For each of 15 random seeds per size class (8–25 nodes), we systematically remove each internal node, triggering cascading path changes. Ground-truth mixed strategies are computed by solving the zero-sum game as a linear program over the post-mobility reward matrix $R'$. Valid examples (those with $|\text{paths}'| > 0$) are serialized to JSONL format, each containing the game topology, removed node identifier, defender/attacker action indices, reward matrix, and oracle strategy vector. Deterministic action canonicalization (lexicographic sorting) ensures identical game states map to identical strategy indices across train/val/test splits, enabling reproducible labels.

\subsection{Splitting Protocol}
To ensure unbiased evaluation, we employ a stratified random split across all 270 network instances.

\begin{itemize}[noitemsep]
    \item \textbf{Stratified Random Split:} Examples are partitioned into train (35\%), validation (15\%), and test (50\%) sets, stratified by graph size so that sizes 8--25 nodes are represented in all splits.

    \item \textbf{Example Generation and Partitioning:} For each graph seed, all
    valid single-node mobility events are enumerated to form the example pool.
    The pooled examples are then partitioned example-wise into train, validation,
    and test sets according to the ratios above, stratified by graph size.

    \item \textbf{Path-Type Stratification:} Test performance is segmented into \textbf{on-path} examples, where the removed node lies on at least one attacker path (560 instances), and \textbf{off-path} examples, where removal does not affect reachability (964 instances). This isolates the effect of topology-altering mobility events.

    \item \textbf{Null Handling:} Examples yielding zero valid attacker paths after removal are excluded from the dataset. Predictions that fail JSON parsing or return strategy $=\texttt{None}$ are flagged as null at evaluation time and excluded from metric computation.
\end{itemize}

The stratified split prioritizes representativeness over size-based extrapolation. Performance is largely invariant to graph size within the 8--25 node range; variation is instead driven by mobility event type and action-space cardinality.

\subsection{Training Process}

The training process involves splitting the dataset, fine-tuning each base model on the equilibrium-prediction task, monitoring validation loss across epochs, and concluding with evaluation against game-theoretic metrics on the held-out test set.

\subsubsection{Dataset Split}

The complete dataset of mobility events was divided into training, validation, and test sets with a split ratio of 35\%, 15\%, and 50\% respectively (1{,}047 / 444 / 1{,}524 examples),
stratified by graph size so that all sizes in the 8--25 node range are represented in every split. The training set is used to fine-tune each model, the validation set to monitor generalization across epochs, and the test set deliberately the largest partition to provide a stable estimate of exploitability against the oracle equilibrium.

\subsubsection{Fine-Tuning}

Fine-tuning casts equilibrium prediction as constrained sequence generation: each training
example pairs a serialized game instance (attack graph, enumerated defender actions, attacker
paths, payoff matrix, and mobility event) with the oracle mixed strategy rendered as a JSON
probability vector. All six models are adapted with LoRA~\cite{hu2022lora} at rank $r{=}8$,
$\alpha{=}16$, and dropout $0.05$, training only the injected low-rank matrices while the base
weights remain frozen. Training runs for 50 epochs at learning rate $2\times10^{-4}$ with the
\texttt{adamw\_8bit} optimizer, weight decay $0.01$, a maximum sequence length of 2048 tokens,
and per-device batch sizes of 1--2 with gradient accumulation to hold the effective batch size
at 8. Fine-tuning is accelerated with Unsloth~\cite{unsloth2023}, and 4-bit quantized base
weights (QLoRA~\cite{dettmers2023qlora}) are used for the memory-constrained
configurations. Target modules are selected per architecture, since MLP projections are named
differently across model families: gated-MLP models (Qwen, Mistral) expose
\texttt{gate\_proj}/\texttt{up\_proj}/\texttt{down\_proj}, LLaMA and Gemma omit
\texttt{gate\_proj}, and Phi uses \texttt{fc1}/\texttt{fc2} in place of the projection pair.
Table~\ref{tab:finetunned_model} reports the resulting adapter sizes.

\begin{table}[hbt!]
\caption{Trainable parameters introduced by LoRA adaptation ($r{=}8$, $\alpha{=}16$).}
\label{tab:finetunned_model}
\scriptsize
\centering
\begin{tabular}{l|r|r|r|r}
\hline
\makecell{Model} & \makecell{Base Model \\ Parameters} & \makecell{LoRA \\ Trainable \\ Parameters} & \makecell{Adapted \\ Layers} & \makecell{Adapter \\ Size} \\
\hline
Phi-3-mini-4k     & 3.8B  & \makecell{6,291,456 \\ (0.165\%)}  & 128 & 25 MB \\
Mistral-7B-v0.3   & 7.2B  & \makecell{20,971,520 \\ (0.289\%)} & 224 & 81 MB \\
Intel Neural-Chat & 7.2B  & \makecell{3,407,872 \\ (0.047\%)}  & 64  & 14 MB \\
Qwen 2.5-7B       & 7.6B  & \makecell{20,185,088 \\ (0.265\%)} & 196 & 78 MB \\
LLaMA-3.1-8B      & 8.0B  & \makecell{16,252,928 \\ (0.202\%)} & 192 & 63 MB \\
Gemma-3-12B       & 12.2B & \makecell{25,362,432 \\ (0.208\%)} & 288 & 97 MB \\
\hline
\end{tabular}
\end{table}

\subsubsection{Testing}

At test time each fine-tuned model is prompted with the serialized post-mobility game instance and emits a JSON probability vector over the enumerated defender action set. Predictions are parsed without repair no truncation, padding, or renormalization is applied so that malformed outputs are surfaced rather than silently corrected. Each prediction is then scored against the oracle equilibrium using the game-theoretic metrics: exploitability gap (regret against the attacker's best response), support F1, total-variation distance, KL divergence, validity rate, and per-example latency. Results are additionally stratified by mobility event type (on-path versus off-path removal) to isolate the effect of topology-altering events.

\subsection{Validation Metrics}
Model predictions are evaluated against ground-truth Nash equilibrium strategies using five complementary metrics:

\begin{itemize}

    \item \textbf{Validity Rate:} Fraction of predictions forming a valid probability distribution over the post-mobility defender action set: the output vector must have length $|\mathcal{A}_d(s')|$, have all entries $\geq -\epsilon$, and satisfy $|\sum_i \hat{x}_i - 1| \leq \epsilon$ with tolerance $\epsilon = 10^{-3}$. Predictions fail validity in three distinct ways unparseable or \texttt{None} output, a well-formed vector of the wrong length, or a parseable vector whose entries do not normalize.
    
    \item \textbf{Support F1 Score:} Harmonic mean of precision and recall for predicted action support (actions with mass $> 0.01$) against oracle support. Measures whether the model identifies which defender actions should receive nonzero probability in the mixed strategy.
    
    \item \textbf{Exploitability Gap:} Regret incurred if an optimal attacker plays best-response against the model's predicted strategy $\hat{\sigma}_d$, relative to the minimax value $V^*$:
    \begin{equation}
    \text{Gap} = \max_{a \in A} \sum_{d \in D} \hat{\sigma}_d \cdot R(d, a) - V^*
    \end{equation}
    Lower gap indicates closer alignment to oracle strategy. Analyzed by game type: on-path instances (where removed node lies on attacker paths) versus off-path instances.
    
    \item \textbf{Total Variation Distance:} Statistical distance between predicted and oracle distributions:
    \begin{equation}
    \text{TV}(\hat{\sigma}_d, \sigma_d^*) = \frac{1}{2}\sum_{i=1}^{|D|} |\hat{\sigma}_{d,i} - \sigma_{d,i}^*|
    \end{equation}
    Bounded in $[0, 1]$; value 0 indicates identical distributions.
    
    \item \textbf{KL Divergence:} Information-theoretic distance measuring distributional mismatch:
    \begin{equation}
    D_{\text{KL}}(\sigma_d^* \| \hat{\sigma}_d) = \sum_i \sigma_d^*(i) \log\left(\frac{\sigma_d^*(i)}{\hat{\sigma}_d(i)}\right)
    \end{equation}
    Penalizes confident mispredictions (high $\hat{\sigma}_d$ where oracle has low mass) more heavily than diffuse errors.
\end{itemize}

\section{Experimental Results}
\label{sec:results}

We evaluate four classes of predictor on the held-out test set of $1{,}524$ mobility events ($560$ on-path, $964$ off-path removals) drawn from graphs of $8$--$25$ nodes. The first class comprises three non-learned baselines: uniform allocation over the post-mobility defender action set, a greedy heuristic that concentrates honeypots on structurally central edges, and a Mock-Oracle that re-solves the linear program of Eq.~\eqref{eq:lp} at each state and therefore bounds achievable performance. The second class consists of six open-weight instruction-tuned models (Qwen 2.5-7B, LLaMA-3.1-8B, Phi-3-mini-4k, Mistral-7B-v0.3, Intel Neural-Chat-7B, and Gemma-3-12B) evaluated zero-shot, with no gradient updates. The third class comprises the same six architectures after LoRA fine-tuning on the training split, using the per-architecture configurations of Table~\ref{tab:finetunned_model}. The fourth class provides an external reference point: four proprietary models accessed through their respective APIs (GPT-5 Mini, GPT-5 Nano, Claude Haiku 4.5, and Gemini 3.5 Flash-Lite), likewise evaluated zero-shot. 
%The fine-tuned model is publicly accessible on Huggingface \footnote{\scriptsize{\url{https://huggingface.co/AbuSayed1/DGA-Meta-Llama-3-8B-FineTunnedModel}}} and our GitHub page \footnote{\scriptsize{\url{https://github.com/MDABUSAYED/StratosphereLinuxIPS/tree/LLM/modules/flowalerts/LLM}}}.

\begin{figure*}[t]
  \centering
  \includegraphics[width=\textwidth,height=10cm]{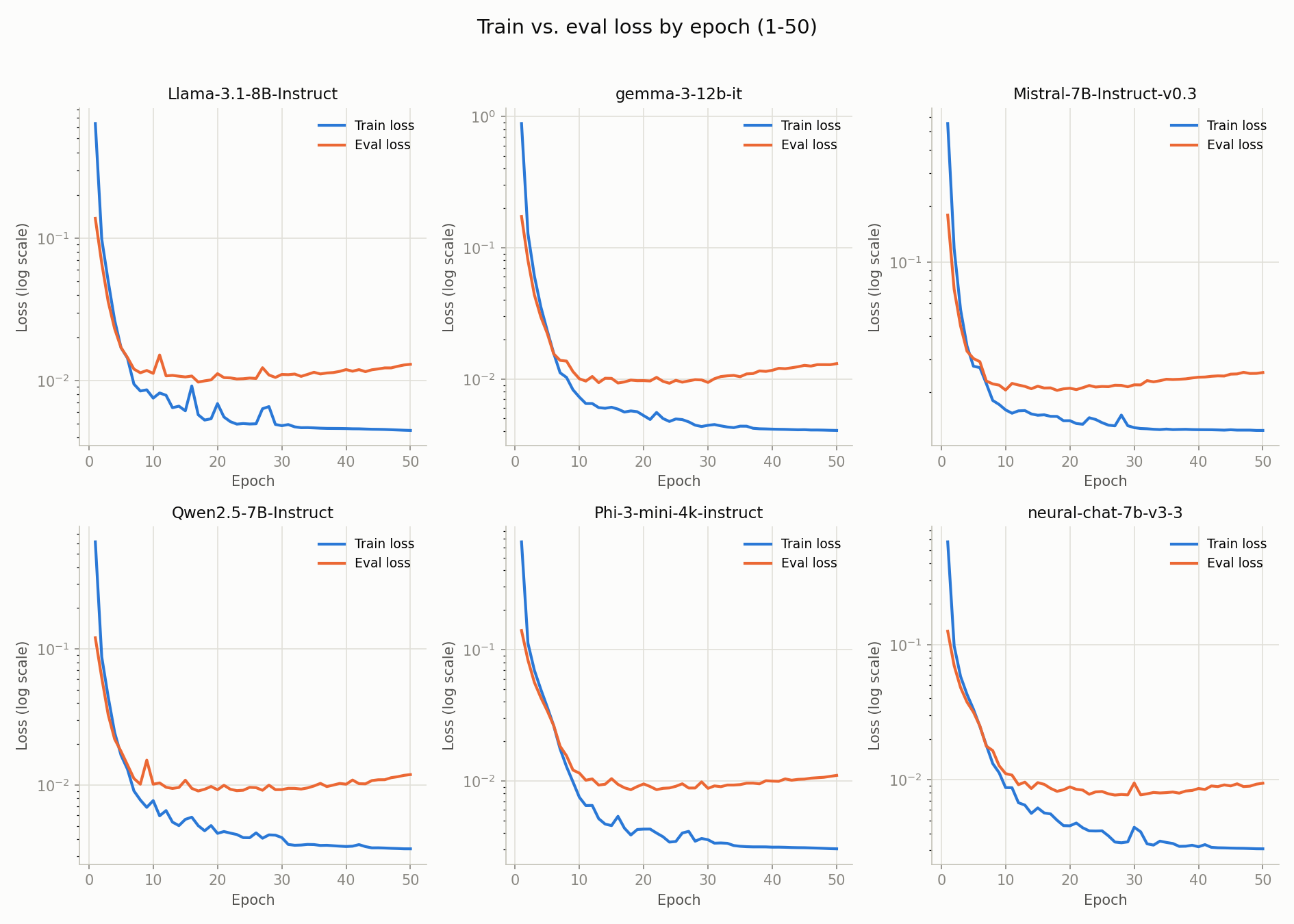}
  \caption{Train vs.\ evaluation loss by epoch for the fine-tuned models.}
  \label{fig:loss_curves}
\end{figure*}

All predictors receive the identical serialized game instance and emit a
probability vector over the same enumerated defender action set, so that the four classes share a single input/output contract and are directly comparable. Outputs are parsed without repair, and every prediction is scored against the oracle equilibrium using the metrics of Section~\ref{sec:dataset}: exploitability gap, achieved defender value, support F1, total-variation distance, KL divergence, validity rate, and per-example latency. Results are additionally stratified by mobility type to separate removals that perturb the equilibrium from those that do not. Table~\ref{tab:agent_performance} reports aggregate and stratified performance for all agents; Fig.~\ref{fig:loss_curves} reports the corresponding training dynamics for the fine-tuned models.

\textbf{Loss Dynamics and the Limits of Validation Loss.}
Figure~\ref{fig:loss_curves} plots training and validation loss over the full 50-epoch schedule for all six fine-tuned architectures. Every model exhibits the same three-phase behavior: a steep decline over the first 5--10 epochs as the output schema is acquired, a plateau in which validation loss reaches its minimum, and a divergence phase in which training loss continues to fall while validation loss rises. The minima occur at epoch 17 for Qwen ($0.0091$) and LLaMA ($0.0098$), epoch 18 for Phi-3-mini ($0.0086$) and Mistral ($0.0205$), epoch 24 for Gemma ($0.0093$), and epoch 27 for Intel Neural-Chat ($0.0077$). By epoch 50, validation loss has risen $22.4\%$--$41.1\%$ above these minima while
training loss has fallen a further $12.7\%$--$39.3\%$. Read conventionally, this is textbook overfitting and would mandate early stopping between epochs 17 and 27.

Task-level evaluation contradicts that reading. As reported above, checkpoints selected by validation loss exhibited substantially worse test-set exploitability than the epoch-50 models, motivating our decision to train the full schedule. The discrepancy arises because validation loss measures token-level cross-entropy over the serialized JSON target, whereas exploitability measures regret against an optimal attacker best-response. These objectives are only loosely coupled: a prediction that reproduces the oracle's numerals imprecisely may still place mass on the correct support and remain difficult to exploit, while a prediction with lower token loss may concentrate mass incorrectly.

The effect is more pronounced across models than across checkpoints. Ranking the six architectures by terminal validation loss yields the order Intel Neural-Chat ($0.0094$), Phi-3-mini ($0.0110$), Qwen ($0.0120$), LLaMA ($0.0130$), Gemma ($0.0131$), Mistral ($0.0255$); ranking the same models by exploitability gap in Table~\ref{tab:agent_performance} yields almost the reverse: Mistral ($216.80$), LLaMA ($257.43$), Qwen ($329.96$), Gemma ($355.42$), Intel Neural-Chat ($452.38$), Phi-3-mini ($522.30$). The two orderings are
negatively rank-correlated ($\rho_s = -0.77$), and the extremes invert outright: the model with the highest validation loss achieves the lowest exploitability gap, while the model with the lowest validation loss ranks fifth of six on the task metric. With $n=6$ this correlation does not attain statistical significance ($p > 0.05$), so we present it as indicative rather than established. It nonetheless suggests that validation loss is unreliable both for checkpoint selection within a model and for ranking models against one another on this task.
Where a verifiable game-theoretic criterion is available, it should supersede the surrogate loss for model selection.

%\subsection{Results and Analysis}

\begin{table*}[h!]
\centering
\caption{Performance on Honeypot Allocation Under Network Mobility}
\label{tab:agent_performance}
\normalsize
\resizebox{\textwidth}{!}{%
\begin{tabular}{@{}l|cccccccc|cccc@{}}
\toprule
\textbf{Strategy/Model} & \textbf{Gap} & \textbf{Value} & \textbf{F1 Score} & \textbf{TV Dist.} & \textbf{KL Div.} & \textbf{Valid.} & \textbf{Latency (s)} & \textbf{On-Path Gap} & \textbf{Off-Path Gap} & \textbf{Ratio} & \textbf{N (On/Off)} \\
\midrule
\multicolumn{12}{c}{\textit{Baselines}} \\
Uniform Distribution & 479.94 & -843.91 & 0.637 & 0.650 & 9.58 & 1.00 & 3.73e-06 & 450.94 & 496.79 & 0.91 & 560 / 964 \\
Greedy Centrality & 636.30 & -1000.26 & 0.381 & 0.695 & 5.99 & 1.00 & 3.14e-05 & 562.85 & 678.96 & 0.83 & 560 / 964 \\
Mock-Oracle (LP) & 9.09 & -373.06 & 0.842 & 0.029 & 0.013 & 1.00 & 0.100 & 6.63 & 10.52 & 0.63 & 560 / 964 \\
\midrule
\multicolumn{12}{c}{\textit{Zero-Shot LLMs: Open-Weight Models}} \\
Qwen 2.5-7B           & 590.02 & -953.99 & 0.428 & 0.751 & 12.64 & 0.10 & 2.64  & 516.50 & 632.73 & 0.82 & 560 / 964 \\
LLaMA-3.1-8B          & 587.53 & -951.50 & 0.364 & 0.805 & 13.05 & 0.35 & 7.39  & 508.32 & 633.55 & 0.80 & 560 / 964 \\
Phi-3-mini-4k         & 646.29 & -1010.26 & 0.426 & 115.30 & 12.54   & 0.10 & 5.81 & 518.32 & 720.63  & 0.72 & 560 / 964 \\
Mistral-7B-v0.3       & 597.61 & -961.58 & 0.462 & 0.778 & 12.75 & 0.12 & 5.83  & 532.34 & 635.53 & 0.84 & 560 / 964 \\
Intel Neural-Chat-7B  & 867.21 & -1231.18 & 0.439 & 1.109  & 12.42 & 0.09 & 5.77 & 495.61 & 1083.08 & 0.46 & 560 / 964 \\
%OpenChat-3.5-7B       & 599.61 & -963.58 & 0.438 & 0.742 & 12.07 & 1.00 & 5.76  & 536.45 & 636.30 & 0.84 & 560 / 964 \\
Gemma-3-12B           & 603.70 & -967.67 & 0.533 & 0.774 & 11.72 & 0.28 & 10.38 & 481.74 & 674.55 & 0.71 & 560 / 964 \\
\midrule
\multicolumn{12}{c}{\textit{Fine-Tuned LLMs: Open-Weight Models}} \\
Qwen 2.5-7B (FT)      & 329.96 & -693.93 & 0.718 & 0.406 & 5.92  & 0.61 & 2.56  & 512.49 & 223.93 & 2.29 & 560 / 964 \\
LLaMA-3.1-8B (FT)     & 257.43 & -621.40 & 0.802 & 0.283 & 3.97  & 0.72 & 2.07  & 479.56 & 128.40 & 3.74 & 560 / 964 \\
Phi-3-mini-4k (FT)    & 522.30 & -886.26 & 0.523 & 1.281 & 9.34 & 0.31 & 2.23 & 585.28 & 485.71 & 1.21 & 560 / 964 \\
Mistral-7B-v0.3 (FT)  & 216.80 & -580.76 & 0.818 & 0.398 & 3.55  & 0.72 & 2.80  & 432.96 & 91.22  & 4.75 & 560 / 964 \\
Intel Neural-Chat (FT)& 452.38 & -816.34 & 0.614 & 0.533 & 7.05 & 0.42 & 2.51 & 540.08 & 401.43 & 1.35 & 560 / 964 \\
Gemma-3-12B (FT)      & 355.42 & -719.39 & 0.682 & 1.443 & 5.68  & 0.58 & 20.52 & 509.74 & 265.77 & 1.92 & 560 / 964 \\
\midrule
\multicolumn{12}{c}{\textit{Zero-Shot LLMs: Proprietary Models (Cost-Efficient Tier)}} \\
GPT-5 Mini            & 517.67 & -881.64 & 0.535 & 0.726 & 11.30 & 0.52 & 3.33 & 487.67 & 535.10 & 0.91 & 560 / 964 \\
GPT-5 Nano            & 523.47 & -887.44 & 0.515 & 0.727 & 11.59 & 0.43 & 1.61 & 508.79 & 532.00 & 0.96 & 560 / 964 \\
Claude Haiku 4.5      & 540.78 & -904.75 & 0.474 & 0.672 & 10.44 & 0.62 & 2.97 & 579.03 & 518.57 & 1.12 & 560 / 964 \\
Gemini 3.5 Flash-Lite & 511.16 & -875.12 & 0.514 & 0.707 & 11.16 & 0.57 & 1.31 & 531.39 & 499.40 & 1.06 & 560 / 964 \\
\bottomrule
\end{tabular}%
}
\end{table*}

\textbf{Baselines Establish the Reference Frame.}
The three baselines bracket the achievable range. The Mock-Oracle, which re-solves the linear program of Eq.~\eqref{eq:lp} at each state, attains a gap of $9.09$ and support F1 of $0.842$; its residual nonzero gap reflects solver tolerance rather than a modeling error, and it serves as the empirical ceiling. Uniform allocation yields a gap of $479.94$, while Greedy Centrality --- despite using structural information --- is substantially worse at $636.30$. That a naive uniform mixture
outperforms a centrality heuristic by $156$ gap units is itself instructive: in a zero-sum setting, concentrating honeypots on structurally central edges produces a predictable defense that an optimal attacker routes around, whereas uniform randomization is at least unexploitable in expectation. This makes Uniform, not Greedy, the meaningful floor against which learned predictors must be judged.

\textbf{Zero-Shot Models Do Not Reach the Uniform Floor.}
No zero-shot model, open-weight or proprietary, achieves a gap below Uniform's $479.94$. Open-weight models span $587.53$ (LLaMA-3.1-8B) to $867.21$ (Intel Neural-Chat-7B), and the proprietary cost-efficient tier spans $511.16$ (Gemini 3.5 Flash-Lite) to $540.78$ (Claude Haiku 4.5). Uniform also attains a higher support F1 ($0.637$) than every zero-shot model except none, exceeding even the strongest zero-shot F1 of $0.535$. The proprietary models do form a distinct and consistently stronger band: all four outperform all six open-weight models on gap, and achieve validity rates of $0.43$--$0.62$ against $0.09$--$0.35$ for open-weight models. Model scale and instruction-tuning quality therefore improve output discipline and reduce exploitability, but not enough to clear a baseline that requires no reasoning at all.

\textbf{Fine-Tuning Is the Dominant Effect.}
Supervised fine-tuning reduces the exploitability gap for every architecture, by $19.2\%$ (Phi-3-mini) to $63.7\%$ (Mistral-7B), with a mean reduction of approximately $45\%$. Five of six fine-tuned models fall below the Uniform floor, and the two strongest Mistral-7B ($216.80$) and LLaMA-3.1-8B ($257.43$) approach less than half of it. The gains are consistent across metrics rather than confined to exploitability: Mistral improves support F1 from $0.462$ to $0.818$, TV distance from $0.778$ to $0.398$, and KL divergence from $12.75$ to $3.55$. Critically, the fine-tuned open-weight models also surpass the proprietary zero-shot tier: Mistral's gap of $216.80$ is less than half that of the best proprietary model, despite operating at $7$B parameters. Task-specific adaptation of a small open-weight model thus yields more than model scale does, which directly supports the amortization argument of Section~\ref{sec:reformulation}: the cost of fine-tuning is paid once, offline, and recovers performance unavailable to any zero-shot predictor at any tier. Phi-3-mini is the sole exception, remaining above Uniform at $522.30$; we attribute this to its smaller capacity ($3.8$B) combined with the lowest validity rate among fine-tuned models ($0.31$).

\textbf{Mobility Type Reverses Direction Under Fine-Tuning.}
The on-path/off-path stratification exposes a qualitative difference that aggregate metrics conceal. The Mock-Oracle finds off-path removals easier in absolute terms but exhibits a ratio of $0.63$, and every zero-shot open-weight model likewise reports ratios below unity ($0.46$--$0.84$), meaning they perform \emph{worse} on off-path events those in which the removed node lies on no attack path and the equilibrium is therefore least perturbed. This is the opposite of the expected ordering and indicates that these models are not conditioning on path relevance at all; their errors scale with the size of the residual action space rather than with strategic difficulty. Every fine-tuned model inverts this pattern, with ratios of $1.21$--$4.75$. Mistral's off-path gap collapses to $91.22$ and LLaMA's to $128.40$, within an order of magnitude of the oracle's $10.52$, while their on-path gaps remain at $432.96$ and $479.56$. Fine-tuning therefore does not merely improve average accuracy: it teaches the model which mobility events actually alter the equilibrium, recovering a structural distinction that is absent from zero-shot behavior. Among zero-shot models, only Claude Haiku ($1.12$) and Gemini Flash-Lite ($1.06$) show the correct ordering, suggesting limited emergent sensitivity to path structure at the proprietary tier.

\textbf{Validity Remains the Binding Constraint.}
Validity rates reveal that a substantial fraction of predictions are not usable probability distributions at all. Open-weight zero-shot models produce valid simplices in as few as $9\%$ of cases; fine-tuning raises this to $0.31$--$0.72$ but does not solve it. Four entries report TV distances above unity (Phi-3-mini zero-shot at $115.30$, Intel Neural-Chat zero-shot at $1.109$, and the fine-tuned Phi-3-mini and Gemma-3-12B at $1.281$ and $1.443$), which is impossible for a metric bounded in $[0,1]$ and indicates predictions whose entries are neither normalized nor dimensionally consistent with the action set. Because we score predictions without repair, these failures surface in the metrics rather than being masked by post-hoc normalization. The practical implication is that constrained decoding or an explicit verification step, rather than additional fine-tuning data, is the most direct route to further improvement.

\textbf{Latency.}
All LLM predictors are slower than the solver they aim to replace. Inference latencies range from $1.31$\,s (Gemini 3.5 Flash-Lite) to $20.52$\,s (fine-tuned Gemma-3-12B, which requires sequential decoding for stability at $12$B parameters), against $0.100$\,s for the Mock-Oracle. Fine-tuning reduces latency relative to zero-shot inference for most architectures LLaMA drops from $7.39$\,s to $2.07$\,s  because fine-tuned models emit terse JSON rather than discursive reasoning. At the graph sizes evaluated here ($n \leq 25$, $H=1$), the amortization argument therefore does not yet yield a wall-clock advantage; the analysis of Section~\ref{sec:compute_comparison} shows why this ordering is specific to small instances and reverses as $H$ grows.

\section{Conclusion}
\label{sec:conclusion}

This work asked whether a fine-tuned language model can amortize the cost of repeatedly re-solving a security game whose structure changes under network mobility. We formalized the re-solve loop as a mapping from a graph and a mobility event to an equilibrium mixed strategy, cast its approximation as an amortized optimization problem, and built a solver-grounded corpus that pairs post-mobility game instances with linear-programming equilibrium labels. Six open-weight architectures were fine-tuned with LoRA under architecture-specific configurations and evaluated against non-learned baselines, their own zero-shot counterparts, and a proprietary reference tier, using verifiable game-theoretic metrics rather than subjective or model-judged criteria.

Three findings emerge. First, task adaptation dominates model scale: fine-tuning reduces exploitability for every architecture, and the strongest fine-tuned open-weight models outperform all zero-shot predictors, including proprietary ones, while no zero-shot model reaches the uniform-allocation floor. Second, fine-tuning does more than improve calibration stratifying by mobility type reveals that zero-shot models fail to distinguish removals that perturb the equilibrium from those that do not, a distinction every fine-tuned model recovers. Third, the conventional training signal is unreliable here: validation loss reaches its minimum
well before the checkpoints that perform best on the task, and across architectures it is negatively rank-correlated with exploitability, indicating that a verifiable game-theoretic criterion should supersede surrogate loss for model selection wherever one is available.

\bibliographystyle{IEEEtran}
\bibliography{references}

\end{document}